\documentclass{article}
\usepackage{ijcai22}

\usepackage{times}

\usepackage{soul}
\usepackage{url}
\usepackage[hidelinks]{hyperref}
\usepackage[utf8]{inputenc}
\usepackage[small]{caption}
\usepackage{graphicx}
\usepackage{amsmath}
\usepackage{booktabs}
\usepackage{algorithm}
\usepackage{algorithmic}
\usepackage{graphicx}
\usepackage{booktabs}       % professional-quality tables
\usepackage{xcolor}         % colors
\usepackage{color}
\usepackage[noabbrev,nameinlink,capitalise]{cleveref}
\usepackage[aboveskip=2pt,belowskip=2pt]{subcaption}
\usepackage{comment}
\usepackage{multirow,setspace}
\usepackage{adjustbox}
\usepackage{amsfonts}       % blackboard math symbols
\usepackage{amsmath,amssymb}
\usepackage{amsthm,latexsym}
\usepackage{bm,amsbsy,isomath}
\usepackage{enumitem}
\usepackage{titlesec}
\usepackage{relsize}
\usepackage[normalem]{ulem}
\usepackage{xfrac}
\usepackage{nicefrac}       % compact symbols for 1/2, etc.
\usepackage{etoolbox}
\usepackage{microtype}      % microtypography
\usepackage{stmaryrd}

% ...
\usepackage{xspace}
\usepackage{adjustbox}         % better centering
\usepackage{multirow,siunitx}
\usepackage{caption}
\usepackage{comment}

\usepackage{pgf}
\usepackage{tikz}
\usepackage{forest}

\usetikzlibrary{arrows,decorations.pathmorphing,decorations.footprints,fadings,calc,trees,mindmap,shadows,decorations.text,patterns,positioning,shapes,matrix,fit}
\usetikzlibrary{graphs}%,graphdrawing,quotes}
\usetikzlibrary{decorations.pathreplacing}
\usetikzlibrary{arrows,shadows,backgrounds}
\usetikzlibrary{arrows.meta}
\usetikzlibrary{positioning,fit}
\usetikzlibrary{automata}
\usetikzlibrary{matrix}
\usetikzlibrary{karnaugh}
\usetikzlibrary{shapes.symbols,shapes.misc,shapes.arrows}
\usetikzlibrary{matrix,chains,scopes,decorations.pathmorphing}
\newtheorem{definition}{Definition}

\newtheorem{example}{Example}

\crefname{theorem}{Theorem}{Theorems}
\crefname{lemma}{Lemma}{Lemmas}
\crefname{proposition}{Proposition}{Propositions}
\crefname{definition}{Definition}{Definitions}
\crefname{corollary}{Corollary}{Corollaries}
\crefname{example}{Example}{Examples}
\crefname{claim}{Claim}{Claims}
\crefname{assumption}{Assumption}{Assumptions}

\newcommand{\fml}[1]{{\mathcal{#1}}}

\newcommand{\tn}[1]{\textnormal{#1}}

\newcommand{\mbf}[1]{\ensuremath\mathbf{#1}}

\newcommand{\mbb}[1]{\ensuremath\mathbb{#1}}
\newcommand{\mrm}[1]{\ensuremath\mathrm{#1}}

\newcommand{\Prob}{\ensuremath\tn{Pr}}
\newcommand{\prob}{\Prob}

\newcommand{\bfdelta}{\ensuremath\pmb{\delta}}

\newcommand{\waxp}{\ensuremath\mathsf{WeakAXp}}

\newcommand{\axp}{\ensuremath\mathsf{AXp}}

\newcommand{\wdrset}{\ensuremath\mathsf{WeakPAXp}}
\newcommand{\drset}{\ensuremath\mathsf{PAXp}}
\newcommand{\mdrset}{\ensuremath\mathsf{MinPAXp}}
\newcommand{\adrset}{\ensuremath\mathsf{ApproxPAXp}}

\newcommand{\ite}{\tn{ite}}

\DeclareMathOperator*{\limply}{\rightarrow}

\definecolor{gray}{rgb}{.4,.4,.4}
\definecolor{midgrey}{rgb}{0.5,0.5,0.5}
\definecolor{middarkgrey}{rgb}{0.35,0.35,0.35}
\definecolor{darkgrey}{rgb}{0.3,0.3,0.3}
\definecolor{darkred}{rgb}{0.7,0.1,0.1}
\definecolor{midblue}{rgb}{0.2,0.2,0.7}
\definecolor{darkblue}{rgb}{0.1,0.1,0.5}
\definecolor{darkgreen}{rgb}{0.1,0.5,0.1}
\definecolor{defseagreen}{cmyk}{0.69,0,0.50,0}

\newcommand{\jnoteF}[1]{}

\newcolumntype{L}[1]{>{\raggedright\let\newline\\\arraybackslash\hspace{0pt}}m{#1}}
\newcolumntype{C}[1]{>{\centering\let\newline\\\arraybackslash\hspace{0pt}}m{#1}}
\newcolumntype{R}[1]{>{\raggedleft\let\newline\\\arraybackslash\hspace{0pt}}m{#1}}

\tikzset{
  0 my edge/.style={densely dashed, my edge},
  my edge/.style={-{Stealth[]}},
}

\renewcommand{\baselinestretch}{0.96875} %0.9575

\renewcommand{\dblfloatpagefraction}{0.999}
\renewcommand{\dbltopfraction}{0.999}
\setlist{nosep}

\providetoggle{long}
\settoggle{long}{false}

\title{Provably Precise, Succinct and Efficient Explanations for Decision Trees}

\author{
Yacine Izza$^{1,2}$\footnote{Contact Author}\and
Alexey Ignatiev$^2$\and
Nina Narodytska$^3$\and
Martin C. Cooper$^4$\And
Joao Marques-Silva$^5$\\
\affiliations
$^1$University of Toulouse, Toulouse, France\\
$^2$Monash University, Melbourne, Australia\\
$^3$VMware Research, CA, USA\\
$^4$University of  Toulouse \MakeUppercase{\romannumeral 3}, IRIT, Toulouse, France\\
$^5$IRIT, CNRS, Toulouse, France\\
\emails
\{yacine.izza, alexey.ignatiev\}@monash.edu,
nnarodytska@vmware.com,
martin.cooper@irit.fr,
joao.marques-silva@irit.fr
}

\begin{document}

\maketitle

\begin{abstract}
  Decision trees (DTs) embody interpretable classifiers. DTs have been
  advocated for deployment in high-risk applications, but also for
  explaining other complex classifiers.
  Nevertheless, recent work has demonstrated that predictions in DTs
  ought to be explained with rigorous approaches.
  Although rigorous explanations can be computed in polynomial time
  for DTs, their size may be beyond the cognitive limits of human
  decision makers.
  This paper investigates the computation of $\delta$-relevant sets
  for DTs.
  $\delta$-relevant sets denote explanations that are succinct and
  provably precise. These sets represent generalizations of rigorous
  explanations, which are precise with probability one, and so they
  enable trading off explanation size for precision.
  %
  %This paper investigates the computation of $\delta$-relevant sets,
  %which represent explanations for DTs
  %%which are not only provably precise but also
  %that are succinct and provably precise, trading off explanation size
  %for precision.
  %
  The  paper proposes two logic encodings for computing
  smallest $\delta$-relevant sets for DTs. The paper further devises a 
  polynomial-time algorithm for computing $\delta$-relevant
  sets which are not guaranteed to be subset-minimal, but for which
  the experiments show to be most often subset-minimal in practice.
  %shown to be most often subset-minimal in practice.
  %%  or close to being being subset-minimal
  %
  The experimental results
  also
  demonstrate the practical efficiency of
  computing smallest $\delta$-relevant sets.
  %The results also
  %confirm that approximate of subset-minimal $\delta$-relevant sets
  %that are computed in polynomial time are in general subset-minimal
  %(even if not guaranteed to be).
  %
\end{abstract}

\section{Introduction} \label{sec:intro}

Decision trees (DTs) are widely regarded as epitomizing interpretable
classifiers in machine learning
(ML)~\cite{breiman-ss01,freitas-sigkdd13,molnar-bk20}. In a DT, a
prediction is associated with a concrete tree path. Such a tree path is
expected to be succinct given the number of features, and this is
expected to be the case for sparse DTs.
Moreover, recent work advocates using DTs (and other interpretable
models) in high-risk applications~\cite{rudin-naturemi19}%
\footnote{%
  The European Union (EU) has recently identified a number of examples
  of high-risk applications~\cite{eu-aiact21}.
},
with the main justification being the interpretability of DTs. The
perceived importance of DTs has also motivated a growing body of work
on learning provably optimal and/or sparse
DTs~\cite{bertsimas-ml17,rudin-nips19,schaus-cj20,szeider-aaai21a}.
DTs have also been proposed for explaining more complex models,
including LIME~\cite{guestrin-kdd16} or SHAP~\cite{lundberg-nips17},
again due to being interpretable.
Nevertheless, the interpretability of DTs has been disputed by recent
work~\cite{barcelo-nips20,marquis-kr21,hiims-kr21}, concretely when
interpretability equates with succinctness of explanations. The main
observation, supported by experimental evidence, is that paths in DTs
may not be as succinct as rigorously computed explanations. In turn,
this implies that predictions of DT classifiers should also be
explained.
Although computing smallest rigorous explanations for DTs has been
shown to be NP-hard~\cite{barcelo-nips20}, it is also the case that a
number of relevant queries on DTs has been shown to be
tractable~\cite{marquis-kr21},
including the computation of subset-minimal rigorous
explanations~\cite{hiims-kr21}. As a consequence, the computation of
rigorous explanations for DTs was shown to be feasible in practice.
Nevertheless, as shown by the experiments in this paper, explanations
for DTs can also be too large, exceeding the cognitive limits of
human decision makers~\cite{miller-pr56}.
One possible alternative is the computation of $\delta$-relevant sets,
i.e.\ approximate explanations that offer strong probabilistic
guarantees in terms of precision (i.e.\ measuring how good the
explanation actually is given the function computed by the classifier).
However, recent work~\cite{kutyniok-jair21} proved that, for a general
class of classifiers, computing such approximate explanations is hard
for $\tn{NP}^{\tn{PP}}$. This complexity result hints at the practical
infeasibility of exactly computing approximate explanations that offer 
probabilistic guarantees in terms of precision.
This paper shows that the problem is computationally easier in the
concrete case of DTs, in theory and in practice.

This paper investigates the computation of approximate explanations
that offer probabilistic guarantees in terms of precision,
specifically for the concrete case of DTs.
First, the paper shows that approximate explanations, which offer strong
probabilistic guarantees but which are not guaranteed to be
subset-minimal, can be computed in polynomial time. The paper then
shows that the decision problem for approximate explanations with a
size bound is in NP. This result involves two encodings of the problem
of computing a smallest approximate explanation into Satisfiability
Modulo Theories (SMT), one involving non-linear arithmetic, and
another involving linear arithmetic.

The experimental results demonstrate that computing smallest
approximate explanations can be solved efficiently for large size
DTs. More importantly, the experimental results suggest that, most
often, the polynomial-time algorithm for computing approximate
explanations yields explanations that are indeed subset-minimal, with
negligible running times.
The experimental results also compare the algorithms proposed in this
paper, with the well-known model-agnostic explainer
Anchor~\cite{guestrin-aaai18}.
The difference in the quality of computed explanations is conclusive,
%and the poor results obtained with Anchor,
further validating prior evidence that model-agnostic explainers offer
poor guarantees on the precision of computed explanations. 

\section{Preliminaries} \label{secLprelim}

\paragraph{Classification problems.}
This paper considers classification problems, which are defined on a
set of features (or attributes) $\fml{F}=\{1,\ldots,m\}$ and a set of
classes $\fml{K}=\{c_1,c_2,\ldots,c_K\}$.
Each feature $i\in\fml{F}$ takes values from a domain $\mbb{D}_i$.
In general, domains can be categorical or ordinal, with values that
can be boolean, integer or real-valued.
%, but in this paper we restrict $\mbb{D}_i=\{0,1\}$ and
%$\fml{K}=\{0,1\}$.
%
%
Feature space is defined as
$\mbb{F}=\mbb{D}_1\times{\mbb{D}_2}\times\ldots\times{\mbb{D}_m}$;
% $=\{0,1\}^{m}$
$|\mbb{F}|$ represents the total number of points in $\mbb{F}$.
For boolean domains, $\mbb{D}_i=\{0,1\}=\mbb{B}$, $i=1,\ldots,m$, and
$\mbb{F}=\mbb{B}^{m}$.
The notation $\mbf{x}=(x_1,\ldots,x_m)$ denotes an arbitrary point in
feature space, where each $x_i$ is a variable taking values from
$\mbb{D}_i$. The set of variables associated with features is
$X=\{x_1,\ldots,x_m\}$.
Moreover, the notation $\mbf{v}=(v_1,\ldots,v_m)$ represents a
specific point in feature space, where each $v_i$ is a constant
representing one concrete value from $\mbb{D}_i$. %$=\{0,1\}$.
An ML classifier $\mbb{M}$ is characterized by a (non-constant)
\emph{classification function} $\kappa$ that maps feature space
$\mbb{F}$ into the set of classes $\fml{K}$,
i.e.\ $\kappa:\mbb{F}\to\fml{K}$.
An \emph{instance} %(or example)
denotes a pair $(\mbf{v}, c)$, where $\mbf{v}\in\mbb{F}$ and
$c\in\fml{K}$, with $c=\kappa(\mbf{v})$. 
%(We also use the term \emph{instance} to refer to $\mbf{v}$, leaving
%$c$ implicit.)

\paragraph{Decision trees.}
A decision tree $\fml{T}=(V,E)$ is a directed acyclic graph,
with $V=\{1,\ldots,|V|\}$, having at most one path between every pair
of nodes. $\fml{T}$ has a root node, characterized by having no
incoming edges. All other nodes have one incoming edge. We consider
univariate decision trees where each non-terminal node is associated
with a single feature $x_i$.
Each edge is labeled with a literal, relating a feature (associated
with the edge's starting node) with some values (or range of values)
from the feature's domain. We will consider literals to be of the form
$x_i\in\mbb{E}_i$. $x_i$ is a variable that denotes the value taken
by feature $i$, whereas $\mbb{E}_i\subseteq\mbb{D}_i$ is a subset of
the domain of feature $i\in\fml{F}$.
The type of literals used to label the edges of a DT allows the
representation of the DTs generated by a wide range of decision tree
learners (e.g.~\cite{utgoff-ml97}). 
The set of paths of $\fml{T}$ is denoted by $\fml{R}$.
$\mrm{\Phi}(R_k)$ denotes the set of features associated with path
$R_k\in\fml{R}$, one per node in the tree, with repetitions allowed.
It is assumed that for any $\mbf{v}\in\mbb{F}$ there exists
\emph{exactly} one path in $\fml{T}$ that is consistent with
$\mbf{v}$. By \emph{consistent} we mean that the literals associated
with the path are satisfied (or consistent) with the feature values
in~$\mbf{v}$.
%%\\ % This *must* be here!!! ???
%
%\begin{comment}
%\end{comment}

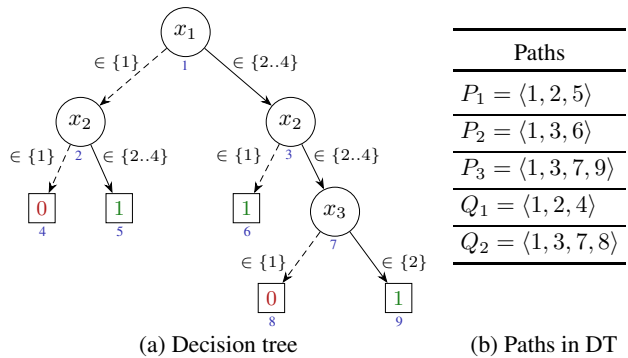
\begin{figure}[t]
  \begin{subfigure}[b]{0.68875\columnwidth}
    \hspace*{-0.395cm}
    \scalebox{0.885}{\renewcommand{\oplus}{1}
\renewcommand{\ominus}{0}
\tikzset{every label/.style={xshift=-0.35ex,
  yshift=-6.225ex,
  text width=1ex,
  align=right, inner sep=1pt, font=\tiny, text=midblue}}
\tikzset{tlabel/.style={xshift=0.25ex, yshift=2ex, text width=1ex,
    align=right, inner sep=1pt, font=\tiny, text=midblue}}
\forestset{
  BDT/.style={
    for tree={
      %l=1.575cm,s sep=1.5cm,
      l=1.35cm,s sep=1.5cm,
      if n children=0{}{circle},
      draw,
      edge={
        my edge
      },
      if n=1{
        edge+={0 my edge},
      }{},
    }
  },
}
\begin{forest}
  BDT
  [$x_{1}$, label={1}
    [$x_{2}$, s sep=0.75cm, label={2},
      edge label={node[near start,left,xshift=-0.75pt] {{\scriptsize$\in\{1\}$}}}
      [{\footnotesize\color{darkred}$\ominus$},
        label={[xshift=0.25ex,yshift=1.875ex]4},
        edge label={node[near start,left,xshift=-0.5pt]
          {{\scriptsize$\in\{1\}$}}}]
      [{\footnotesize\color{darkgreen}$\oplus$},
        label={[xshift=0.25ex,yshift=1.875ex]5},
        edge label={node[near start,right,xshift=-1pt]
          {{\scriptsize$\in\{2..4\}$}}}]
    ]
    [$x_2$, s sep=0.75cm, label={3},
      edge label={node[near start,right,xshift=1pt]
        {{\scriptsize$\in\{2..4\}$}}}
      [{\footnotesize\color{darkgreen}$\oplus$},
        label={[xshift=0.25ex,yshift=1.875ex]6},
        edge label={node[near start,left,xshift=-1pt]
          {{\scriptsize$\in\{1\}$}}}]
      [$x_3$, label={7},
        edge label={node[near start,right,xshift=-1pt]
          {{\scriptsize$\in\{2..4\}$}}}
        [{\footnotesize\color{darkred}$\ominus$},
          label={[xshift=0.25ex,yshift=1.875ex]8},
          edge label={node[pos=0.6,left,xshift=-1pt]
            {{\scriptsize$\in\{1\}$}}}]
        [{\footnotesize\color{darkgreen}$\oplus$},
          label={[xshift=0.25ex,yshift=1.875ex]9},
          edge label={node[pos=0.6,right,xshift=0.25pt]
            {{\scriptsize$\in\{2\}$}}}]
      ]
    ]
  ]
\end{forest}}
    \caption{Decision tree} \label{fig:runex:dt}
  \end{subfigure}
  \begin{subfigure}[b]{0.3\columnwidth}
    \renewcommand{\arraystretch}{1.25}
    \renewcommand{\tabcolsep}{3pt}
    \scalebox{0.895}{
      \begin{tabular}{l} \toprule
        \multicolumn{1}{c}{Paths}
        \\ \toprule
        $P_1=\langle1,2,5\rangle$
        \\[1.5pt] \hline
        $P_2=\langle1,3,6\rangle$
        \\[1.5pt] \hline
        $P_3=\langle1,3,7,9\rangle$
        \\[1.5pt] \hline
        $Q_1=\langle1,2,4\rangle$
        \\[1.5pt] \hline
        $Q_2=\langle1,3,7,8\rangle$
        \\ \bottomrule
      \end{tabular}
    }

    \bigskip\bigskip%\medskip

    \caption{Paths in DT} \label{fig:runex:paths}
  \end{subfigure}
  \caption{Example DT} \label{fig:runex}
\end{figure}

\paragraph{Running example.}
\cref{fig:runex} shows the example DT used throughout the paper. This
example DT also illustrates the notation used to represent DTs.
The set of paths $\fml{R}$ is partitioned into two sets $\fml{P}$ and
$\fml{Q}$, such that the paths in $\fml{P}=\{P_1,P_2,P_3\}$
yield a prediction of 1, and such that the paths in
$\fml{Q}=\{Q_1,Q_2\}$ yield a prediction of 0. (In general, $\fml{P}$
denotes the paths with prediction $c$, and $\fml{Q}$ denotes the paths
with prediction other than $c$, in $\fml{K}\setminus\{c\}$.)

\paragraph{Formal explanations.}
%\subparagraph*{Abductive and constrastive explanations.}
%
We now define formal explanations. In contrast with the well-known
model-agnostic approaches to
XAI~\cite{guestrin-kdd16,lundberg-nips17,guestrin-aaai18,pedreschi-acmcs19},
formal explanations are model-precise, i.e.\ their definition reflects
the model's computed function.
Prime implicant (PI) explanations~\cite{darwiche-ijcai18} denote a
minimal set of literals (relating a feature value $x_i$ and a constant
$v_i\in\mbb{D}_i$) %from its domain $\mbb{D}_i$)
that are sufficient for the prediction. PI-explanations are related
with abduction, and so are also referred to as abductive explanations
($\axp$)~\cite{inms-aaai19}.
More recently, PI-explanations have been studied
%from a knowledge compilation
%perspective~\cite{marquis-kr20,marquis-kr21}, but also
in terms of their computational
complexity~\cite{barcelo-nips20,marquis-kr21}.
Recent work on formal explanations
includes for
example~\cite{kwiatkowska-ijcai21,ims-ijcai21,mazure-cikm21}. %,rubin-aaai22
Formally, given $\mbf{v}=(v_1,\ldots,v_m)\in\mbb{F}$ with
$\kappa(\mbf{v})=c$, an $\axp$ is any minimal subset
$\fml{X}\subseteq\fml{F}$ such that, 
\begin{equation} \label{eq:axp}
  \forall(\mbf{x}\in\mbb{F}).
  \left[
    \bigwedge\nolimits_{i\in{\fml{X}}}(x_i=v_i)
    \right]
  \limply(\kappa(\mbf{x})=c)
\end{equation}
i.e.\ the features in $\fml{X}$ are sufficient for the prediction
when these take the values dictated by $\mbf{v}$, and $\fml{X}$ is
irreducible. Also, a non-minimal set such that ~\eqref{eq:axp} holds
is a $\waxp$.
%%\label{page:eq:axp}
%
$\axp$'s can be viewed as answering a `Why?' question, i.e.\ why is some 
prediction made given some point in feature space.
%
%A different view of explanations is a contrastive
%explanation~\cite{miller-aij19}, which answers a `Why Not?' question,
%i.e.\ which features can be changed to change the prediction.
%
%This paper addresses exclusively $\axp$'s.
Contrastive explanations~\cite{miller-aij19} offer a different view of
explanations, but these are beyond the scope of the paper.
%
%
\begin{comment}
%
A formal definition of contrastive explanation 
is proposed in recent work~\cite{inams-aiia20}.
%
Given $\mbf{v}=(v_1,\ldots,v_m)\in\mbb{F}$ with $\kappa(\mbf{v})=c$, a
contrastive explanation (CXp) is any minimal subset
$\fml{Y}\subseteq\fml{F}$ such that,
%
\begin{equation} \label{eq:cxp}
  \exists(\mbf{x}\in\mbb{F}).\bigwedge\nolimits_{j\in\fml{F}\setminus\fml{Y}}(x_j=v_j)\land(\kappa(\mbf{x})\not=c) %\not\in\fml{Y}
\end{equation}
%
Building on the results of R.~Reiter in model-based
diagnosis~\cite{reiter-aij87},~\cite{inams-aiia20} proves a minimal
hitting set (MHS, or hypergraph transversal~\cite{berge-bk84}) duality
relation between AXps and CXps, i.e.\ AXps are MHSes of CXps and
vice-versa.
%
Throughout the paper, $\tn{(M)HS}(\mbb{Z})$ %(resp.~$\tn{MHS}(\mbb{Z})$)
denote the set of (minimal) hitting sets of $\mbb{Z}$.
%
\end{comment}

\begin{example}
  The computation of $\waxp$'s and $\axp$'s is illustrated with the
  DT from~\cref{fig:runex}. The instance considered throughout is
  $\mbf{v}=(v_1,v_2,v_3)=(4,4,2)$, with $c=\kappa(\mbf{v})=\mbf{1}$.
  The point $\mbf{v}$ is consistent with $P_3$, and
  $\mrm{\Phi}(P_3)=\{1,2,3\}$.
  \cref{tab:cprob} (columns 1 to 4) analyzes three sets of features
  $\{1,2,3\}$, $\{1,3\}$ and $\{3\}$ in terms of being a $\waxp$ or an 
  $\axp$. The decision on whether each set is a $\waxp$ or an 
  $\axp$ can be obtained by analyzing all the 32 points in feature
  space, or by using an off-the-shelf algorithm. (Analysis of
    all points in feature space is ommited for brevity.)
\end{example}

\begin{table*}[t]
  \centering
  \renewcommand{\arraystretch}{1.05}
  \renewcommand{\tabcolsep}{0.325em}
  \scalebox{0.875}{
    \begin{tabular}{c|c||c|c||c|c|c||c|c|c|c|c|c} \toprule
      $\fml{S}$ & $\fml{U}$ &
      $\waxp$? & $\axp$? & 
      $\prob_{\mbf{x}}(\kappa(\mbf{x})=c|(\mbf{x}_{\fml{S}}=\mbf{v}_{\fml{S}}))$ &
      $\wdrset$? & $\drset$? &
      $\#(\fml{S})$ & $\#(P_1)$ & $\#(P_2)$ & $\#(P_3)$ & $\#(Q_1)$ & $\#(Q_2)$
      \\ \toprule
      $\{1,2,3\}$ & $\emptyset$ &
      Yes & No &
      $1\ge\delta$ &
      Yes & No &
      1 & 0 & 0 & 1 & 0 & 0
      \\ \midrule
      $\{1,3\}$ & $\{2\}$ &
      Yes & Yes &
      $1\ge\delta$ &
      Yes & No &
      4 & 0 & 1 & 3 & 0 & 0
      \\ \midrule
      $\{3\}$ & $\{1,2\}$ &
      No & -- &
      $\sfrac{15}{16}=0.9375\ge\delta$ &
      Yes & Yes &
      16 & 3 & 3 & 9 & 1 & 0
      \\ %\midrule
      \bottomrule
    \end{tabular}
  }
  \caption{Examples of sets of fixed features given $\mbf{v}=(4,4,2)$
    and $\delta=0.93$}
  \label{tab:cprob}
\end{table*}

\paragraph{Relevant sets.}
%\paragraph{$\delta$-relevant sets.}
%
$\delta$-relevant sets were proposed in more recent
work~\cite{kutyniok-jair21} as a generalized formalization of 
explanations. $\delta$-relevant sets can be viewed as
\emph{probabilistic} PIs, %~\cite{},
with $\axp$'s representing 1-relevant sets,
%a special case of $\delta$-relevant sets,
i.e.\ probabilistic PIs that are actual PIs. We briefly overview the
definitions related with relevant sets. 
%
%We consider a generalized definition of
%min-$\delta$-relevant-input~\cite{kutyniok-jair21}.
%
The assumptions regarding the probabilities of logical propositions
are those made in earlier work~\cite{kutyniok-jair21}.
Let $\prob_{\mbf{x}}(A(\mbf{x}))$ denote the probability of some
proposition $A$ defined on the vector of variables
$\mbf{x}=(x_1,\ldots,x_m)$, i.e.
\begin{equation} \label{eq:pdefs}
  \begin{array}{rcl}
    \prob_{\mbf{x}}(A(\mbf{x})) & = &
    \frac{|\{\mbf{x}\in\mbb{F}:A(\mbf{x})=1\}|}{|\{\mbf{x}\in\mbb{F}\}|}
    \\[4.0pt]
    \prob_{\mbf{x}}(A(\mbf{x})\,|\,B(\mbf{x})) & = &
    \frac{|\{\mbf{x}\in\mbb{F}:A(\mbf{x})=1\land{B(\mbf{x})=1}\}|}{|\{\mbf{x}\in\mbb{F}:B(\mbf{x})=1\}|}
  \end{array}
\end{equation}
(Similar to earlier work, it is assumed that the features are
independent and uniformly distributed~\cite{kutyniok-jair21}.
Moreover, the definitions above can be adapted in case some of the
features are real-valued. To keep the notation simple, the paper
studies only categorical and integer-valued features.)

\begin{definition}[$\delta$-relevant set~\cite{kutyniok-jair21}]
  Consider $\kappa:\mbb{B}^{m}\to\fml{K}=\mbb{B}$, $\mbf{v}\in\mbb{B}^m$,
  $\kappa(\mbf{v})=c\in\mbb{B}$, and
  $\delta\in[0,1]$. $\fml{S}\subseteq\fml{F}$ is a $\delta$-relevant
  set for $\kappa$ and $\mbf{v}$ if,
  \begin{equation} \label{eq:drs}
    \prob_{\mbf{x}}(\kappa(\mbf{x})=c\,|\,\mbf{x}_{\fml{S}}=\mbf{v}_{\fml{S}})\ge\delta
  \end{equation}
  (where the restriction of $\mbf{x}$ to the variables with indices in
  $\fml{S}$ is represented by
  $\mbf{x}_{\fml{S}}=(x_i)_{i\in\fml{S}}$).
\end{definition}
(Observe that
$\prob_{\mbf{x}}(\kappa(\mbf{x})=c\,|\,\mbf{x}_{\fml{S}}=\mbf{v}_{\fml{S}})$ 
is often referred to as the \emph{precision} of
$\fml{S}$~\cite{guestrin-aaai18,nsmims-sat19}.)
Thus, a $\delta$-relevant set represents a set of features which, if
fixed to some pre-defined value (taken from a reference vector
$\mbf{v}$), ensure that the probability of the prediction being the
same as the one for $\mbf{v}$ is no less than $\delta$. 

\begin{definition}[Min-$\delta$-relevant set] \label{def:mdrs}
  Given $\kappa$, $\mbf{v}\in\mbb{B}^{m}$, and $\delta\in[0,1]$, find
  the smallest $k$, such that there exists $\fml{S}\subseteq\fml{F}$, with
  $|\fml{S}|={k}$, and $\fml{S}$ is a $\delta$-relevant set for
  $\kappa$ and $\mbf{v}$.
\end{definition}
With the goal of proving the computational complexity of finding a
minimum-size set of features that is a $\delta$-relevant set, earlier
work~\cite{kutyniok-jair21} restricted the definition to the case
where $\kappa$ is represented as a boolean circuit.

\iftoggle{long}{%
  (Boolean circuits were restricted to propositional formulas defined 
  using the operators $\lor$, $\land$ and $\neg$, and using a set of
  variables representing the inputs; this explains the choice of
  \emph{inputs} over \emph{sets} in earlier
  work~\cite{kutyniok-jair21}.)%
}{}

%% Decide whether to include related work...

%\subsection{Related Work} \label{ssec:relw}
\paragraph{Related work.~}
%
%$\delta$-relevant sets~\cite{kutyniok-jair21} generalize
%PI-explanations~\cite{darwiche-ijcai18}.
Besides $\delta$-relevant sets,
there is work that also offers
%%have been other attempts at learning implicit DTs that also offer
strong probabilistic guarantees by implicitly learning
DTs~\cite{tan-nips21}. As argued in this and earlier
work~\cite{barcelo-nips20,hiims-kr21,marquis-kr21}, learning a DT may
not suffice to reveal logically sound explanations.
%
%Another line of
There is work~\cite{vandenbroeck-ijcai21} that can be related with
$\delta$-relevant sets, but it opts instead for exploiting heuristic
methods for computing explanations.

\section{$\bfdelta$-Relevant Sets for Decision Trees} \label{sec:rsdt}

%\jnote{%
%  ...
%}

Observe that~\cref{def:mdrs} imposes no restriction on the
representation of the classifier that is assumed in earlier
work~\cite{kutyniok-jair21}, i.e.\ the logical representation of
$\kappa$ need not be a boolean circuit.
As a result, we extend~\cref{def:mdrs},
%the definitions from earlier work~\cite{kutyniok-jair21},
as detailed below.

\subsection{Generalizations}
%\paragraph{Generalizations.}
%
A \emph{weak probabilistic $\axp$} ($\wdrset$) is a pick of fixed
features for which the conditional probability of predicting the
correct class $c$ exceeds $\delta$, given $c=\kappa(\mbf{v})$. (The
classifier is only required to compute function $\kappa$). Thus,
$\fml{S}\subseteq\fml{F}$ is a  $\wdrset$ if,
% $\delta$-relevant set
%
\begin{align} \label{eq:wcdrs2}
  \wdrset&(\fml{S};\mbb{F},\kappa,\mbf{v},\delta)  %\:\, := %\:\:
  \nonumber \\
  :=\,\: & \prob_{\mbf{x}}(\kappa(\mbf{x})=c\,|\,\mbf{x}_{\fml{S}}=\mbf{v}_{\fml{S}})
  \ge \delta
  \\[1.0pt]
  :=\,\: &\frac{%
    |\{\mbf{x}\in\mbb{F}:\kappa(\mbf{x})=c\land(\mbf{x}_{\fml{S}}=\mbf{v}_{\fml{S}})\}|
  }{%
    |\{\mbf{x}\in\mbb{F}:(\mbf{x}_{\fml{S}}=\mbf{v}_{\fml{S}})\}|
  }
  \ge\delta \nonumber
\end{align}
which means that the fraction of the number of models predicting the
target class and consistent with the fixed features (represented by
$\fml{S}$), given the total number of points in feature space
consistent with the fixed features, must exceed $\delta$.
(Observe that the difference to~\eqref{eq:drs} is solely that features
and classes are no longer required to be boolean.)
Moreover, a \emph{probabilistic $\axp$} ($\drset$) $\fml{X}$ is a
$\wdrset$ that is also subset-minimal, 
%%An actual DRSet is a subset-minimal $\wdrset$,
% $\delta$-relevant set
\begin{align}
  \drset&(\fml{X};\mbb{F},\kappa,\mbf{v},\delta) \::= \nonumber \\
  &\wdrset(\fml{X};\mbb{F},\kappa,\mbf{v},\delta) \:\:\land \\
  &\forall(\fml{X}'\subsetneq\fml{X}). %
  \neg\wdrset(\fml{X}';\mbb{F},\kappa,\mbf{v},\delta) \nonumber
\end{align}
Minimum-size $\drset$'s ($\mdrset$) generalize Min-$\delta$-relevant
sets in~\cref{def:mdrs}. 

\begin{example}
  \cref{tab:cprob} summarizes the values of
  $\prob_{\mbf{x}}(\kappa(\mbf{x})=c)|(\mbf{x}_{\fml{S}}=\mbf{v}_{\fml{S}})$
  (column 5) for the sets $\{1,2,3\}$, $\{1,3\}$ and $\{3\}$.The table
  also includes information on whether each set is a $\wdrset$ or a 
  $\drset$ (columns 6 and 7).
\end{example}

We recall that the goal of our work is to efficiently compute precise
and succinct explanations. However, in case of probabilistic $\axp$'s a
critical observation is that $\wdrset$ is a \emph{non-monotone
  predicate}; hence standard algorithms for computing a subset-minimal
set are \emph{not} guaranteed to yield subset-minimal
sets~\cite{msjm-aij17}.
The results in~\cref{sec:res} validate this observation.

\begin{comment}
%
%
\paragraph{{\boldmath$\axp$}'a and {\boldmath$\drset$}'s for running example.}
%
The computation of $\axp$'s and $\drset$'s is illustrated with the DT
from~\cref{fig:runex}.

\begin{example}
  For the DT of \cref{fig:runex}, the instance considered throughout is
  $\mbf{v}=(v_1,v_2,v_3)=(4,4,2)$, with $c=\kappa(\mbf{v})=\mbf{1}$.
  The point $\mbf{v}$ is consistent with $P_3$. The goal is to compute a
  $\delta$-relevant set (DRSet(!)) given $\delta=0.93$.
  %
  Let $\#(R_k)$ denote the number of points in feature space that are
  consistent with path $R_k$. Moreover, let $\#(\fml{X})$ denote the
  total number of points in feature space that are consistent with the
  set of \emph{fixed} features $\fml{X}\in\fml{F}$.
  %
  \cref{tab:cprob} summarizes the computation of
  $\prob_{\mbf{x}}(\kappa(\mbf{x})=c)|(\mbf{x}_{\fml{S}}=\mbf{v}_{\fml{S}})$
  for different sets $\fml{S}$. The table also includes information on
  whether each set is a $\waxp$, $\axp$, $\wdrset$ or $\drset$.
  %
  The set $\{1,3\}$ represents an AXp, since for any point consistent
  with the assignments $x_1=4$ and $x_3=2$, the prediction is 1.
  %
  %
  However, by setting $\fml{S}=\{3\}$, the probability of predicting
  $\mbf{1}$ given a point consistent with $x_3=2$ still exceeds
  $\delta$, since $\sfrac{15}{16}=93.75\%$. Hence, $\{3\}$ is a DRSet(!)
  for $\mbf{v}=(4,4,2)$ when $\delta=0.93$.
\end{example}
%
%
\end{comment}

\subsection{Computing Path Probabilities} \label{sec:pps}

% 1. 
This section investigates how to compute, in the case of DTs, the
conditional probability,
\begin{equation} \label{eq:cprob}
  \prob_{\mbf{x}}(\kappa(\mbf{x})=c\,|\,\mbf{x}_{\fml{X}}=\mbf{v}_{\fml{X}})
\end{equation}
where $\fml{X}$ is a set of fixed features (whereas the other features
are not fixed, being deemed universal), and $(\mbf{v},c)$ is an
instance. (Also, note that \eqref{eq:cprob} is the left-hand side 
of~\eqref{eq:drs}).
To motivate the proposed approach, let us first analyze how can we
compute $\prob_{\mbf{x}}(\kappa(\mbf{x})=c)$, where
$\fml{P}\subseteq\fml{R}$ is the set of paths in the DT with
prediction $c$.
Let $\mrm{\Lambda}(R_k)$ denote the set of literals (each of the form
$x_i\in\mbb{E}_i$) in path $R_k\in\fml{R}$. If a feature $i$ is tested
multiple times along path $R_k$, then $\mbb{E}_i$ is the intersection
of the sets in each of the literals on $i$.
The number of values of $\mbb{D}_i$ consistent with literal
$x_i\in\mbb{E}_i$ is $|\mbb{E}_i|$.
Finally, the features \emph{not} tested along $R_k$ are denoted by
$\mrm{\Psi}(R_k)$.
For path $R_k$, the probability that a randomly chosen point in
feature space is consistent with $R_k$ (i.e.\ the \emph{path
  probability} of $R_k$) is given by,
\[
\prob(R_k) =
\nicefrac{\left[\prod_{(x_i\in\mbb{E}_i)\in\mrm{\Lambda}(R_k)}|\mbb{E}_i|
    \times\prod_{i\in\mrm{\Psi}(R_k)}|\mbb{D}_i|\right]}{|\mbb{F}|}
\]
As a result, we get that,
\[
\prob_{\mbf{x}}(\kappa(\mbf{x})=c)={\textstyle\sum\nolimits}_{R_k\in\fml{P}}\prob(R_k)
\]

Given an instance $(\mbf{v},c)$ and a set of fixed features $\fml{X}$
(and so a set of universal features $\fml{F}\setminus\fml{X}$), we now
detail how to compute~\eqref{eq:drs}.
Since some features will now be declared universal, multiples paths
with possibly different conditions can become consistent. For example,
in~\cref{fig:runex} if feature 1 and 2 are declared universal, then
(at least) paths $P_1$, $P_2$ and $Q_1$ are consistent with some of
the possible assignments.
Although universal variables might seem to complicate the computation
of the conditional probability, this is not the case.

A key observation is that the feature values that make a path
consistent are disjoint from the values that make other paths
consistent. This observation allows us to compute the models 
consistent with each path and, as a result, to
compute~\eqref{eq:drs}.
Let $n_{ik}$ represent the (integer) number of assignments to feature
$i$ that are consistent with path $R_k\in\fml{R}$, given
$\mbf{v}\in\mbb{F}$ and $\fml{X}\subseteq\fml{F}$.
The value of $n_{ik}$ is defined as follows:
\begin{enumerate}[nosep]
\item If $i$ is fixed:
  \begin{enumerate}[nosep]
  \item If $i$ is tested along $R_k$ and the value of $x_i$ is
    inconsistent with $R_k$, i.e.\ there exists a literal
    $x_i\in\mbb{E}_i\in\mrm{\Lambda}(R_k)$ and
    $\{v_i\}\cap{E}_i=\emptyset$, then $n_{ik}=0$;
  \item If $i$ is tested along $R_k$ and the value of $x_i$ is
    consistent with $R_k$, then $n_{ik}=1$;
  \item If $i$ is not tested along $R_k$, then $n_{ik}=1$.
  \end{enumerate}
\item Otherwise, $i$ is universal:
  \begin{enumerate}[nosep]
  \item If $i$ is tested along $R_k$, with some literal
    $x_i\in\mbb{E}_i$, then $n_{ik}=|\mbb{E}_i|$;
  \item If $i$ is not tested along $R_k$, then $n_{ik}=|\mbb{D}_i|$.
  \end{enumerate}
\end{enumerate}
Using the definition of $n_{ik}$, we can then compute the number of 
assignments consistent with $R_k$ as follows:
\begin{equation}
  \#(R_k;\mbf{v},\fml{X})={\textstyle\prod\nolimits}_{i\in\fml{F}}n_{ik}
\end{equation}
Finally,~\eqref{eq:cprob} is given by,
\begin{align}
\prob_{\mbf{x}}(\kappa(\mbf{x})=c\,|&\,\mbf{x}_{\fml{X}}=\mbf{v}_{\fml{X}})=
\nonumber\\
&\nicefrac%
    {\sum_{P_k\in\fml{P}}\#(P_k;\fml{F}\setminus\fml{X},\mbf{v})}
    {\sum_{R_k\in\fml{R}}\#(R_k;\fml{F}\setminus\fml{X},\mbf{v})}
\end{align}
As can be concluded for the case of a DT,
both
$\prob_{\mbf{x}}(\kappa(\mbf{x})=c\,|\,\mbf{x}_{\fml{X}}=\mbf{v}_{\fml{X}})$
and $\wdrset(\fml{X};\mbb{F},\kappa,\mbf{v},\delta)$ are computed in
polynomial time on the size of the DT.

\begin{example}
  With respect to the DT in~\cref{fig:runex}, and given the instance
  $((4,4,2),1)$, the number of models for each path is shown
  in~\cref{tab:cprob}.
  For example, for set $\{3\}$, we immediately get that
  $\prob_{\mbf{x}}(\kappa(\mbf{x})=c\,|\,\mbf{x}_{\fml{X}}=\mbf{v}_{\fml{X}})=
  \nicefrac{15}{(15+1)}=\nicefrac{15}{16}$.
\end{example}

\subsection{Refining Weak $\bfdelta$-Relevant Sets} \label{sec:adrset}

Recent work showed that, for DTs, one $\axp$ can be computed in
polynomial time~\cite{hiims-kr21}.
A simple polynomial-time algorithm can be summarized as follows.
The $\axp$ $\fml{X}$ is initialized to all the features in $\fml{F}$.
Pick the path consistent with a given instance $(\mbf{v},c)$.
The features not in the path are removed from $\fml{X}$. Then,
iteratively check whether $\fml{X}\setminus\{i\}$ guarantees that all
paths to a prediction in $\fml{K}\setminus\{c\}$ are still
inconsistent. If so, then update $\fml{X}$.
We can use a similar approach for computing one
\emph{approximate} probabilistic $\axp$:  
an $\adrset$
$\fml{X}$ is a $\wdrset$ such that the removal of any \emph{single}
feature $i$ from $\fml{X}$ will falsify
$\wdrset(\fml{X}\setminus\{i\};\mbb{F},\kappa,c,\delta)$
(see~\eqref{eq:wcdrs2}).
To compute an $\adrset$, we start from $\fml{F}$ and remove features
while it is safe to do so, i.e.\ while \eqref{eq:wcdrs2} holds for the
resulting set. The algorithm runs in polynomial time.
Thus, by construction, the resulting $\adrset$ is an over-approximation
of a $\drset$.
However, due to non-monotoniticy of $\wdrset$, a $\adrset$ may not be
subset-minimal and so not a $\drset$.
%
%
%Use the algorithm outlined in the preceding paragraph, but compute the
%precision and check that it is above the target threshold precision
%$\delta$.
%
%Since the precision is always no less than $\delta$, we are guaranteed
%to compute an over-approximate DRSet.
%
%However, because $\wdrset$ is a non-monotone predicate, we cannot
%guarantee that the computed set is subset-minimal.

\subsection{Computing {\boldmath$\mdrset$'s}}  \label{sec:mdrset}

For computing a smallest $\drset$, we propose two SMT encodings, thus
showing that the decision problem is in NP, and that finding a
smallest set requires a logarithmic number of calls to an NP-oracle.
Regarding the two SMT encodings, one involves the multiplication of
integer variables, and so it involves non-linear arithmetic. Given the
structure of the problem, we also show that linear arithmetic can be
used, by proposing a (polynomially) larger encoding.

\paragraph{A multiplication-based SMT encoding.}
Taking into account the definition of path probabilities
(see~\cref{sec:pps}), we now devise a model that computes path
probabilities based on the same ideas.
Let $n_{jk}$ denote the number of elements in $\mbb{D}_j$ consistent
with path $R_k$ (for simplicity, we just use the path index $k$).
If $j$ is not tested along path $R_k$, then if $j$ is fixed, then
$n_{jk}=1$. If not, then $n_{jk}=|\mbb{D}_j|$.
Otherwise, $j$ is tested along path $R_k$.
$n_{jk}$ is 0 if $j$ is fixed (i.e.\ $u_j=0$) and
inconsistent with the values of $\mbb{D}_j$ allowed for path
$R_k$. $n_{jk}$ is 1 if $j$ is fixed and consistent with the values of
$\mbb{D}_j$ allowed for path $R_k$.
If $j$ is not fixed (i.e.\ it is universal), the $n_{jk}$ denotes the
number of domain values of $j$ consistent with path $R_k$.
Let the fixed value of $n_{jk}$ be
$n_{0jk}$ and the \emph{universal} value of $n_{jk}$ be $n_{1jk}$.
Thus, $n_{jk}$ is defined as follows,
\begin{equation}
  n_{jk}=\ite(u_j,n_{1jk},n_{0jk})
\end{equation}
Moreover, let $\eta_k$ denote the number of models of path $R_k$.
Then, $\eta_k$ is defined as follows:
\begin{equation} \label{eq:cntprod}
  %\begin{array}{l}
    \eta_k = {\textstyle\prod\nolimits}_{i\in\mrm{\Phi}(k)}n_{ik}
  %\end{array}
\end{equation}
If the domains are boolean, then we can use a purely boolean
formulation for the problem. However, if the domains are multi-valued,
then we need this formulation.

Recall what we must ensure that~\eqref{eq:wcdrs2} holds true.
In the case of DTs, since we can count the models associated with each
path, depending on which features are fixed or not, then the previous
constraint can be translated to:
\begin{equation} \label{eq:cdrscond}
  {\textstyle\sum\nolimits}_{R_k\in\fml{P}}\eta_k \ge
  \delta \times {\textstyle\sum\nolimits}_{R_k\in\fml{P}}\eta_k
  +
  \delta \times {\textstyle\sum\nolimits}_{R_k\in\fml{Q}}\eta_k
\end{equation}
Recall that $\fml{P}$ are the paths with the matching
prediction, and $\fml{Q}$ are the rest of the paths.
%, and such that $\fml{R}=\fml{P}\cup\fml{Q}$.

Finally, the soft constraints are of the form $(u_i)$, one for each
feature $i\in\fml{F}\setminus\Psi(R_k)$. (For each $i\in\Psi(R_k)$ we
enforce that the feature is universal by adding a hard clause $(u_i)$.)
The solution to the optimization problem will then be a
\emph{smallest} $\wdrset$, and so also a $\drset$. (The minimum-cost
solution is well-known to be computed with a worst-case logarithmic
number of calls (on the number of features) to an SMT solver.) 

%\paragraph{Multiplication-based encoding for the running example.}
%%
%%The simplest encoding represents the problem of computing a smallest
%%DRSet in non-linear arithmetic (which involves the multiplication of
%%multiple integer variables).
%~\\

\begin{table}[t]
  \centering
  \renewcommand{\arraystretch}{1.05}
  \renewcommand{\tabcolsep}{0.425em}
  \scalebox{0.975}{
    \begin{tabular}{c|c|C{1.0cm}C{0.85cm}C{0.85cm}C{0.85cm}C{0.85cm}} \toprule
      Feature & Attr. & $P_1$ & $P_2$ & $P_3$ & $Q_1$ & $Q_2$ \\ \toprule
      \multirow{3}{*}{1}
      & $n_{01k}$ & 0 & 1 & 1 & 0 & 1 \\
      & $n_{11k}$ & 1 & 3 & 3 & 1 & 3 \\ \cline{2-7}
      & $n_{1k}$ &
      \multicolumn{5}{c}{$n_{1k}=\ite(u_1,n_{11k},n_{01k})$}
      \\
      \midrule
      \multirow{3}{*}{2}
      & $n_{02k}$ & 1 & 0 & 1 & 0 & 1 \\
      & $n_{12k}$ & 3 & 1 & 3 & 1 & 3 \\ \cline{2-7}
      & $n_{2k}$ &
      \multicolumn{5}{c}{$n_{2k}=\ite(u_2,n_{12k},n_{02k})$}
      \\
      \midrule
      \multirow{3}{*}{3}
      & $n_{03k}$ & 1 & 1 & 1 & 1 & 0 \\
      & $n_{13k}$ & 2 & 2 & 1 & 2 & 1 \\ \cline{2-7}
      & $n_{3k}$ &
      \multicolumn{5}{c}{$n_{3k}=\ite(u_3,n_{13k},n_{03k})$}
      \\
      \toprule
      \multicolumn{2}{c}{Path counts} &
      \multicolumn{5}{c}{$\eta_{k}=n_{1k}\times{n_{2k}}\times{n_{3k}}$}
      \\
      \bottomrule
    \end{tabular}
  }
  \caption{SMT encoding for multiplication-based encoding}
  \label{tab:smtenc0}
\end{table}

\begin{example}
For the running example, let us consider $\fml{X}=\{3\}$. This means that
$u_1=u_2=1$. As a result, given the instance and the proposed
encoding, we get~\cref{tab:smtenc0} and%
%. In addition, given the set $\fml{X}$, we get
~\cref{tab:runex0}.
\begin{table}[t]
  \centering
  \renewcommand{\arraystretch}{1.0}
  \renewcommand{\tabcolsep}{0.375em}
  \scalebox{0.975}{
    \begin{tabular}{c|ccc|c} \toprule
      Path & $n_{1k}$ & $n_{2k}$ & $n_{3k}$ & $\eta_{k}$ \\ \toprule
      $R_1$ & 1      &  3       &  1       & 3 \\
      $R_2$ & 1      &  3       &  1       & 3 \\
      $R_3$ & 3      &  3       &  1       & 9 \\ \midrule
      $R_4$ & 1      &  1       &  1       & 1 \\
      $R_5$ & 3      &  3       &  0       & 0 \\
      \bottomrule
    \end{tabular}
  }
  \caption{Concrete values for the multiplication-based encoding for
    the case $\fml{X}=\{3\}$, i.e.\ $u_1=u_2=1$ and $u_3=0$}
  \label{tab:runex0}
\end{table}

Finally, by plugging into~\eqref{eq:cdrscond} the values
from~\cref{tab:runex0}, we get: $15 \ge 0.93 \times (15+1)$.
Thus, $\fml{X}$ is a $\wdrset$, and we can show that it is both a
$\drset$ and a $\mdrset$.
Indeed, with $\fml{Y}=\emptyset$, we get
$\prob_{\mbf{x}}(\kappa(\mbf{x})=c\,|\,\mbf{x}_{\fml{Y}}=\mbf{v}_{\fml{Y}})=21/32=0.65625<\delta$. Hence,
$\fml{X}=\{3\}$ is subset-minimal.
Since there can be no $\drset$'s of smaller size, then $\fml{X}$ is
also a $\mdrset$.
%hence $\fml{X}=\{3\}$ is a $\wdrset$ and it is subset-minimal.
\end{example}

\paragraph{An alternative addition-based SMT encoding.}
A possible downside of the SMT encoding described above is the use of
multiplication of variables in~\eqref{eq:cntprod}; this causes the SMT
problem formulation to involve different theories (which may turn out
to be harder to reason about in practice). Given the problem
formulation, we can use an encoding that just uses linear arithmetic.
This encoding is organized as follows.
Let the order of features be: $\langle1,2,\ldots,m\rangle$.
Define $\eta_{j,k}$ as the sum of models of path $R_k$ taking into
account features 1 up to $j$, with $\eta_{0,k}=1$.
Given $\eta_{{j-1},{k}}$, $\eta_{{j},{k}}$ is computed as follows:
\begin{itemize}[nosep]
\item
  Let the domain of feature $j$ be $\mbb{D}_j=\{v_{j1},\ldots,v_{jr}\}$, and
  let $s_{j,l,k}$ denote the number of models taking into account
  features 1 up to $j-1$ and domain values $v_{j1}$ up to
  $v_{j{l-1}}$. Also, let $s_{j,0,k}=0$.
\item For each value $v_{jl}$ in $\mbb{D}_j$, for $l=1,\ldots,r$:
  \begin{itemize}[nosep]
  \item If $j$ is tested along path $R_k$:
    (i) If $v_{jl}$ is inconsistent with path $R_k$, then
    $s_{j,l,k}=s_{j,l-1,k}$;
    (ii) If $v_{jl}$ is consistent with path $R_k$ and with
    $\mbf{v}$, then $s_{j,l,k}=s_{j,l-1,k}+\eta_{{j-1},{k}}$;
    (iii) If $v_{jl}$ is consistent with path $R_k$ but not with
    $\mbf{v}$, or if feature $j$ is not tested in path $R_k$,
    then $s_{j,l,k}=s_{j,l-1,k}+\ite(u_j,\eta_{{j-1},{k}},0)$.
    %\begin{itemize}[nosep]
    %\item If $v_{jl}$ is inconsistent with path $k$, then
    %  $s_{j,l,k}=s_{j,l-1,k}$
    %\item If $v_{jl}$ is consistent with path $k$ and with
    %  $\mbf{v}$, then $s_{j,l,k}=s_{j,l-1,k}+\eta_{{j-1},{k}}$.
    %\item If $v_{jl}$ is consistent with path $k$ but not with
    %  $\mbf{v}$, or if feature $j$ is not tested in path $k$,
    %  then $s_{j,l,k}=s_{j,l-1,k}+\ite(u_j,\eta_{{j-1},{k}},0)$.
    %\end{itemize}
  \item If $j$ is not tested along path $R_k$:
    (i) If $v_{jl}$ is consistent with $\mbf{v}$, then
    $s_{j,l,k}=s_{j,l-1,k}+\eta_{{j-1},{k}}$;
    (ii) Otherwise,
    $s_{j,l,k}=s_{j,l-1,k}+\ite(u_j,\eta_{{j-1},{k}},0)$.
    %\begin{itemize}[nosep]
    %\item If $v_{jl}$ is consistent with $\mbf{v}$, then
    %  $s_{j,l,k}=s_{j,l-1,k}+\eta_{{j-1},{k}}$
    %\item Otherwise,
   %   $s_{j,l,k}=s_{j,l-1,k}+\ite(u_j,\eta_{{j-1},{k}},0)$.
   % \end{itemize}
  \end{itemize}
\item Finally, define $\eta_{{j},{k}}=s_{j,r,k}$.
\end{itemize}

After considering all the features in order, $\eta_{m,k}$
represents the number of models for path $R_k$ given the assigment to 
the $u_j$ variables.
As a result, we can re-write~\eqref{eq:cdrscond} as follows:
\begin{equation} \label{eq:cdrscond2}
  {\textstyle\sum\nolimits}_{R_k\in\fml{P}}\eta_{m,k} \ge
  \delta \times {\textstyle\sum\nolimits}_{R_k\in\fml{P}}\eta_{m,k}
  +
  \delta \times {\textstyle\sum\nolimits}_{R_k\in\fml{Q}}\eta_{m,k}
\end{equation}
As with the multiplication-based encoding, the soft clauses are of the
form $(u_i)$ for $i\in\fml{F}$.

%\begin{example}
%\end{example}
%\input{./tabs/nouse}

\iftoggle{long}{\paragraph{Counting-based SMT encoding for running example.}
Given the proposed counting-based encoding, the resulting encoding is
summarized below.

\begin{table*}[t]
  \centering
  \renewcommand{\arraystretch}{1.15}
  \renewcommand{\tabcolsep}{0.325em}
  \scalebox{0.9575}{
    \begin{tabular}{c|c|c|c|c|c} \toprule
      Var.   &
      $R_1\cong{P_1}$  & $R_2\cong{P_2}$ & $R_3\cong{P_3}$ & $R_4\cong{Q_1}$  & $R_5\cong{Q_2}$ \\
      \toprule
      $s_{1,0,k}$ & $s_{1,0,1}=0$ & $s_{1,0,2}=0$ & $s_{1,0,3}=0$ & $s_{1,0,4}=0$ & $s_{1,0,5}=0$ 
      \\
      $s_{1,1,k}$ &
      $s_{1,0,1}+\ite(u_1,\eta_{0,1},0)$ &
      $s_{1,0,2}$ &
      $s_{1,0,3}$ &
      $s_{1,0,4}+\ite(u_1,\eta_{0,4},0)$ &
      $s_{1,0,5}$ 
      \\ 
      $s_{1,2,k}$ &
      $s_{1,1,1}$ &
      $s_{1,1,2}+\ite(u_1,\eta_{0,2},0)$ &
      $s_{1,1,3}+\ite(u_1,\eta_{0,3},0)$ &
      $s_{1,1,4}$ &
      $s_{1,1,5}+\ite(u_1,\eta_{0,5},0)$
      \\ 
      $s_{1,3,k}$ &
      $s_{1,2,1}$ &
      $s_{1,2,2}+\ite(u_1,\eta_{0,2},0)$ &
      $s_{1,2,3}+\ite(u_1,\eta_{0,3},0)$ &
      $s_{1,2,4}$ &
      $s_{1,2,5}+\ite(u_1,\eta_{0,5},0)$
      \\
      $s_{1,4,k}$ &
      $s_{1,3,1}$ &
      $s_{1,3,2}+\eta_{0,2}$ &
      $s_{1,3,3}+\eta_{0,3}$ &
      $s_{1,3,4}$ &
      $s_{1,3,5}+\eta_{0,5}$
      \\
      \midrule
      $\eta_{1,k}$ & $s_{1,4,1}$ & $s_{1,4,2}$ & $s_{1,4,3}$ & $s_{1,4,4}$ & $s_{1,4,5}$
      \\
      \midrule
      $s_{2,0,k}$ & $s_{2,0,1}=0$ & $s_{2,0,2}=0$ & $s_{2,0,3}=0$ & $s_{2,0,4}=0$ & $s_{2,0,5}=0$
      \\
      $s_{2,1,k}$ &
      $s_{2,0,1}$ &
      $s_{2,0,2}+\eta_{1,2}$ &
      $s_{2,0,3}$ &
      $s_{2,0,4}+\ite(u_2,\eta_{1,4},0)$ &
      $s_{2,0,5}+\ite(u_2,\eta_{1,5},0)$
      \\
      $s_{2,2,k}$ &
      $s_{2,1,1}+\ite(u_2,\eta_{1,1},0)$ &
      $s_{2,1,2}$ &
      $s_{2,1,3}+\ite(u_2,\eta_{1,3},0)$ &
      $s_{2,1,4}$ &
      $s_{2,1,5}$
      \\
      $s_{2,3,k}$ &
      $s_{2,2,1}+\ite(u_2,\eta_{1,1},0)$ &
      $s_{2,2,2}$ &
      $s_{2,2,3}+\ite(u_2,\eta_{1,3},0)$ &
      $s_{2,2,4}$ &
      $s_{2,2,5}$
      \\
      $s_{2,4,k}$ &
      $s_{2,3,1}+\eta_{1,1}$ &
      $s_{2,3,2}$ &
      $s_{2,3,3}+\eta_{1,3},0$ &
      $s_{2,3,4}$ & $s_{2,3,5}$
      \\
      \midrule
      $\eta_{2,k}$ &  $s_{2,4,1}$ & $s_{2,4,2}$ & $s_{2,4,3}$ & $s_{2,4,4}$ & $s_{2,4,5}$
      \\
      \midrule
      $s_{3,0,k}$ & $s_{3,0,1}=0$ & $s_{3,0,2}=0$ & $s_{3,0,3}=0$ & $s_{3,0,4}=0$ & $s_{3,0,5}=0$
      \\
      $s_{3,1,k}$ &
      $s_{3,0,1}+\ite(u_3,\eta_{2,1},0)$ &
      $s_{3,0,2}+\ite(u_3,\eta_{2,2},0)$ &
      $s_{3,0,3}$ &
      $s_{3,0,4}+\ite(u_3,\eta_{2,4},0)$ &
      $s_{3,0,5}+\ite(u_3,\eta_{2,5},0)$
      \\
      $s_{3,2,k}$ &
      $s_{3,1,1}+\eta_{2,1}$ &
      $s_{3,1,2}+\eta_{2,2}$ &
      $s_{3,1,3}+\eta_{2,3}$ &
      $s_{3,1,4}+\eta_{2,4}$ &
      $s_{3,1,5}$
      \\
      \midrule
      $\eta_{3,k}$ & $s_{3,4,1}$ & $s_{3,4,2}$ & $s_{3,4,3}$ & $s_{3,2,4}$ & $s_{3,2,5}$
      \\
      \bottomrule
    \end{tabular}
  }
  \caption{Partial addition-based SMT encoding for paths with
    prediction $\mbf{1}$, with $(\mbf{v},c)=((4,4,2),\mbf{1})$,
    and with $\eta_{0,1}=\eta_{0,2}=\eta_{0,3}=1$} \label{tab:smtenc1}
\end{table*}

\begin{example}
  \cref{tab:smtenc1} summarizes the SMT encoding based on iterated
  summations for paths with either prediction $\mbf{1}$ or $\mbf{0}$.
  %Similarly, \cref{tab:smtenc2} summarizes the SMT encoding based on
  %iterated summations for the paths with prediction $\mbf{0}$.
  The final computed values are then used in the linear
  inequality~\eqref{eq:cdrscond2}, as follows,
  \[
  \eta_{3,1}+\eta_{3,2}+\eta_{3,2}\ge%
  \delta\times(\eta_{3,1}+\eta_{3,2}+\eta_{3,2})+
  \delta\times(\eta_{3,4}+\eta_{3,5})
  \]
  The optimization problem also includes
  $\fml{B}=\{(\neg{u_1}),(\neg{u_2}),(\neg{u_3})\}$ as the soft clauses.
  %
  %\jnote{We might want to check that for $\fml{X}=\{3\}$, we get a
  %  precision that exceeds the threshold.}
  %
  For the counting-based encoding, and from~\cref{tab:smtenc1}, we get
  the values shown in~\cref{tab:runex2}.
  Moreover, we can then confirm that $15\ge0.93\times16$, as intended.
\end{example}

\begin{table}[t]
  \centering
  \renewcommand{\arraystretch}{1.15}
  \renewcommand{\tabcolsep}{0.35em}
  \scalebox{0.9725}{
    \begin{tabular}{c|c|c|c|c|c|} \toprule
      Var.   & $R_1\cong{P_1}$  & $R_2\cong{P_2}$ & $R_3\cong{P_3}$ &
      $R_4\cong{Q_1}$ & $R_5\cong{Q_2}$
      \\
      \toprule
      $s_{1,0,k}$ &
      0 & 0 & 0 & 0 & 0
      \\
      $s_{1,1,k}$ &
      1 &
      0 &
      0 &
      1 &
      0 
      \\ 
      $s_{1,2,k}$ &
      1 &
      2 &
      1 &
      1 &
      1
      \\ 
      $s_{1,3,k}$ &
      1 &
      2 &
      2 &
      1 &
      2
      \\
      $s_{1,4,k}$ &
      1 &
      3 &
      3 &
      1 &
      3
      \\
      \midrule
      $\eta_{1,k}$ & 1 & 3 & 3 & 1 & 3
      \\
      \midrule
      $s_{2,0,k}$ &
      0 & 0 & 0 & 0 & 0
      \\
      $s_{2,1,k}$ &
      0 &
      3 &
      0 &
      0 &
      3
      \\
      $s_{2,2,k}$ &
      1 &
      3 &
      3 &
      1 &
      3
      \\
      $s_{2,3,k}$ &
      2 &
      3 &
      6 &
      1 &
      3
      \\
      $s_{2,4,k}$ &
      3 &
      3 &
      9 &
      1 &
      3
      \\
      \midrule
      $\eta_{2,k}$ & 3 & 3 & 9 & 1 & 3
      \\
      \midrule
      $s_{3,0,k}$ &
      0 & 0 & 0 & 0 & 0
      \\
      $s_{3,1,k}$ &
      0 &
      2 &
      0 &
      0 &
      0
      \\
      $s_{3,2,k}$ &
      3 &
      3 &
      9 &
      1 &
      0
      \\
      \midrule
      $\eta_{3,k}$ & 3 & 3 & 9 & 1 & 0
      \\
      \bottomrule
    \end{tabular}
  }
  \caption{Assignment to variables of addition-based SMT encoding,
    given $\fml{X}=\{3\}$, i.e.\ $u_1=u_2=1$ and $u_3=0$}
  \label{tab:runex2}
\end{table}

}{}

\subsection{Deciding if an {\boldmath$\adrset$} is a {\boldmath$\drset$}}

Deciding whether a set of features $\fml{X}$, representing an
$\adrset$, is subset-minimal can be achieved by using one of the
models above, keeping the features that are already universal, and
checking whether additional universal features can be made to exist.
In addition, we need to add constraints forcing universal features to
remain universal, and at least one of the currently fixed features to
also become universal.
Thus, if $\fml{X}$ is the set of fixed features, the SMT models
proposed in earlier sections is extended with the following
constraints:
\begin{equation} \label{eq:chkdrset}
  {\textstyle\bigwedge\nolimits}_{j\in\fml{F}\setminus\fml{X}}(u_j)
  {\textstyle\bigwedge}
  \left({\textstyle\bigvee\nolimits}_{j\in\fml{X}}u_j\right)
\end{equation}
which allow checking whether some set of fixed features can be
declared universal while respecting the other constraints.
%on precision.

%
\setlength{\tabcolsep}{3.5pt}

\sisetup{%
  math-rm=\textrm
}

\begin{table*}[t]%[h]
\centering
\resizebox{\textwidth}{!}{
  \begin{tabular}{l  S[table-format=3]S[table-format=3]   S[table-format=2]S[table-format=1]S[table-format=2.1]  S[table-format=3]   S[table-format=2]S[table-format=1]S[table-format=2.1]  c  S[table-format=2.2]         S[table-format=2]S[table-format=1]S[table-format=2.1]  c c S[table-format=1.2]     cS[table-format=2]S[table-format=1]S[table-format=2.1]S[table-format=2.1]   c   S[table-format=3.2] }
\toprule[1.2pt]
\multirow{3}{*}{\bf Dataset}  &  \multicolumn{2}{c}{} &   \multicolumn{3}{c}{}  &  & \multicolumn{5}{c}{\bf $\bm\mdrset$} & \multicolumn{6}{c}{$\bm\adrset$}  &  \multicolumn{7}{c}{\bf Anchor} \\
\cmidrule[0.8pt](lr{.75em}){8-12}
\cmidrule[0.8pt](lr{.75em}){13-18}
\cmidrule[0.8pt](lr{.75em}){19-25}
 &   \multicolumn{2}{c}{\bf DT}  &  \multicolumn{3}{c}{\bf Path} &  {$\bm\delta$ } & \multicolumn{3}{c}{\bf Length} & \multicolumn{1}{c}{\bf Prec} &  \multicolumn{1}{c}{\bf Time} &  
\multicolumn{3}{c}{\bf Length} & \multicolumn{1}{c}{\bf Prec} & {\bf m$_{\bm\subseteq}$} & \multicolumn{1}{c}{\bf Time}  &
{\bf D} & \multicolumn{4}{c}{\bf Length} & \multicolumn{1}{c}{\bf Prec} &  \multicolumn{1}{c}{\bf Time}\\  
\cmidrule[0.8pt](lr{.75em}){2-3}
\cmidrule[0.8pt](lr{.75em}){4-6}
\cmidrule[0.8pt](lr{.75em}){8-10}
\cmidrule[0.8pt](lr{.75em}){11-11}
\cmidrule[0.8pt](lr{.75em}){12-12}
\cmidrule[0.8pt](lr{.75em}){13-15}
\cmidrule[0.8pt](lr{.75em}){16-16}
\cmidrule[0.8pt](lr{.75em}){18-18}
\cmidrule[0.8pt](lr{.75em}){20-23}
\cmidrule[0.8pt](lr{.75em}){24-24}
\cmidrule[0.8pt](lr{.75em}){25-25}

& {\bf N}  & {\bf A} &  {\bf M} & {\bf m} &  {\bf avg} &  & {\bf M} & {\bf m }  & {\bf avg } &  {\bf avg } &  {\bf avg }  & 
{\bf M} & {\bf m }  & {\bf avg } &  {\bf avg } & { }  & {\bf avg }  &  
{ } & {\bf M} & {\bf m }  & {\bf avg } & {\bf F$_{\bm\not\in P}$}  & {\bf avg } &  {\bf avg } \\
\toprule[1.2pt]

 &  &  &  &  &  &  100 & 11 & 3 & 6.8 & 100 & 2.34 & 11 & 3 & 6.9 & 100 & 100 & 0.00 & d & 12 & 2 & 7.0 &  26.8 & 76.8 & 0.96 \\
adult & 1241 & 89 & 14 & 3 & 10.7 & 95 & 11 & 3 & 6.2 & 98.4 & 5.36 & 11 & 3 & 6.3 & 98.6 & 99.0 & 0.01 & u & 12 & 3 & 10.0 & 29.4 & 93.7 & 2.20 \\
 &  &  &  &  &   & 90 & 11 & 2 & 5.6 & 94.6 & 4.64 & 11 & 2 & 5.8 & 95.2 & 96.4 & 0.01 &  &  &  &  &  & & \\
\midrule
 &  &  &  &  &  &  100 & 12 & 1 & 4.4 & 100 & 0.35 & 12 & 1 & 4.4 & 100 & 100 & 0.00 & d & 31 & 1 & 4.8 & 58.1 & 32.9 & 3.10 \\
dermatology & 71 & 100 & 13 & 1 & 5.1 & 95 & 12 & 1 & 4.1 & 99.7 & 0.37 & 12 & 1 & 4.1 & 99.7 & 99.3 & 0.00 & u & 34 & 1 & 13.1 & 43.2 & 87.2 & 25.13 \\
 &  &  &  &  &  &  90 & 11 & 1 & 4.0 & 98.8 & 0.35 & 11 & 1 & 4.0 & 98.8 & 100 & 0.00 &  &  &  &  &  &  & \\
\midrule
 &  &  &  &  &  &  100 & 12 & 2 & 4.8 & 100 & 0.93 & 12 & 2 & 4.9 & 100 &  100 & 0.00 & d & 36 & 2 & 7.9 &  44.8 & 69.4 & 1.94 \\
kr-vs-kp & 231 & 100 & 14 & 3 & 6.6 & 95 & 11 & 2 & 3.9 & 98.1 & 0.97 & 11 & 2 & 4.0 & 98.1 & 100 & 0.00 & u & 12 & 2 & 3.6 & 16.6 & 97.3 & 1.81 \\
 &  &  &  &  &  &  90 & 10 & 2 & 3.2 & 95.4 & 0.92 & 10 & 2 & 3.3 & 95.4 & 99.0 & 0.00 &  &  &  &  &  &  & \\
\midrule
 &  &  &  &  &  &  100 & 12 & 4 & 8.2 & 100 & 16.06 & 11 & 4 & 8.2 & 100 &  100 & 0.00 & d & 16 & 3 & 13.2 & 43.1 & 71.3 & 12.22 \\
letter & 3261 & 93 & 14 & 4 & 11.8 & 95 & 12 & 4 & 8.0 & 99.6 & 18.28 & 11 & 4 & 8.0 & 99.5 & 100 & 0.00 & u & 16 & 3 & 13.7 & 47.3 & 66.3 & 10.15 \\
 &  &  &  &  &  &  90 & 12 & 4 & 7.7 & 97.7 & 16.35 & 10 & 4 & 7.8 & 97.8 & 100 & 0.00 &  &  &  &  &  &  & \\
\midrule
 &  &  &  &  &  &  100 & 14 & 3 & 6.4 & 100 & 0.92 & 14 & 3 & 6.5 & 100 &  100 & 0.00 & d & 35 & 2 & 8.6 & 55.4  & 33.6 & 5.43 \\
soybean & 219 & 100 & 16 & 3 & 7.3 & 95 & 14 & 3 & 6.4 & 99.8 & 0.95 & 14 & 3 & 6.4 & 99.8 & 100 & 0.00 & u & 35 & 3 & 19.2 & 66.0 & 75.0 & 38.96 \\
 &  &  &  &  &  &  90 & 14 & 3 & 6.1 & 98.1 & 0.94 & 14 & 3 & 6.1 & 98.2 & 98.5 & 0.00 &  &  &  &  &  &  & \\
\midrule
 &  &  &  &  &  &  0 & 12 & 3 & 7.4 & 100 & 1.23 & 12 & 3 & 7.5 & 100 &  100 & 0.01 & d & 38 & 2 & 6.3 & 65.3 & 63.3 & 24.12 \\
spambase & 141 & 99  & 14 & 3 & 8.5 & 95 & 9 & 1 & 3.7 & 96.1 & 2.16 & 9 & 1 & 3.8 & 96.5 & 100 & 0.01 & u & 57 & 3 & 28.0 & 86.2 & 65.3 & 834.70 \\
 &  &  &  &  &  &  90 & 6 & 1 & 2.4 & 92.4 & 2.15 & 8 & 1 & 2.4 & 92.2 & 100 & 0.01 &  &  &  &  &  &  & \\
\midrule
 &  &  &  &  &  &  100 & 12 & 3 & 6.2 & 100 & 2.01 & 11 & 3 & 6.2 & 100 & 100 & 0.01 & d & 40 & 2 & 16.5 & 80.6 & 32.2 & 532.42 \\
texture & 257 & 100  & 13 & 3 & 6.6 & 95 & 11 & 3 & 5.4 & 99.3 & 2.19 & 11 & 3 & 5.4 & 99.4 & 100 & 0.01 & u & 40 & 5 & 17.5 & 84.4 & 31.6 & 402.07 \\
 &  &  &  &  &  &  90 & 11 & 3 & 5.4 & 98.5 & 2.20 & 11 & 3 & 5.4 & 99.4 & 100 & 0.01 &  &  &  &  &  &  & \\

\bottomrule[1.2pt]
\end{tabular}
}
\caption{%
	Assessing explanations of  $\mdrset$, $\adrset$ and Anchor. 
	(For each dataset, we run the explainers on 500 samples randomly picked 
	from the training data or all training samples if there are less than 500.)	
	In column {\bf DT}, {\bf N} and {\bf A} denote, resp., the number of nodes  
	and the training accuracy of the DT.
	%
	%Column {\bf \#I} denotes the number of tested instances randomly 
	%picked from the training data. 
	%
	Column {$\bm\delta$} reports in (\%) the value of the threshold  $\delta$.
	In column 
	{\bf Path},   {\bf avg} (resp.\ {\bf M} and {\bf m}) denotes the average 
	(resp.\  max. and  min.) depth of paths consistent with the instances. 
 	In column  {\bf Length},  {\bf avg} (resp.\ {\bf M} and {\bf m}) denotes 
	the average  (resp.\  max. and  min.)  length of the explanations; and 
	{\bf F$_{\bm\not\in P}$} denotes the avg.\  \% of features in Anchors 
	that do not belong to the consistent paths. 
	{\bf Precision} reports in  (\%) the average precision (defined in (\ref{eq:drs})) of
	resulting explanations. 
	%
	%{$\bm\drset$} shows the number in  (\%) of $\adrset$'s that are $\drset$'s. 
	{\bf m$_{\bm\subseteq}$}	shows the number in  (\%) of $\adrset$'s that are 
	subset-minimal, i.e.\  $\drset$'s. 
	{\bf Time} reports (in seconds)  the average runtime to compute an explanation.	
	Finally,  {\bf D}  indicates which distribution is applied 
	on data given to Anchor: either  data distribution (denoted by {d}) or 
	uniform distribution (denoted by {u}). 
} \label{tab:res}
\end{table*}

\setlength{\tabcolsep}{4pt}
\let\lpr\undefined
\let\rpr\undefined
\newcommand{\lpr}{(}
\newcommand{\rpr}{)}

\sisetup{%
  math-rm=\textrm
}

\begin{table*}[t]
\centering
\resizebox{0.7\textwidth}{!}{
  \begin{tabular}{l  S[table-format=3] c  S[table-format=3]  ccS[table-format=2.2]    ccS[table-format=1.2]  }
\toprule[1.2pt]
\multirow{2}{*}{\bf Dataset} &   & {\bf Path}  &   & \multicolumn{3}{c}{$\bm\mdrset$} & \multicolumn{3}{c}{$\bm\adrset$} \\
\cmidrule[0.8pt](lr{.75em}){3-3}
\cmidrule[0.8pt](lr{.75em}){5-7}
\cmidrule[0.8pt](lr{.75em}){8-10}
  & {\bf \#I} & {\bf Length} & {$\bm\delta$ } & {\bf Length} & {\bf Precision} & {\bf Time} &  {\bf Length} & {\bf Precision} &{\bf Time} \\  

\toprule[1.2pt]

 &  &  & 100 & 8.2 $\pm$ 0.5 & 100 & 5.84 & 8.2 $\pm$ 0.5 & 100 & 1.24  \\
adult & 160 & 12.3 $\pm$ 1.4 & 95 & 7.3 $\pm$ 1.1 & 97.7 & 35.19 & 7.5 $\pm$ 0.9 & 98.0 & 0.01  \\
 &  &  & 90 & 6.4 $\pm$ 1.4 & 93.5 & 35.23 & 6.8 $\pm$ 1.1 & 94.3 & 0.01  \\
\midrule
 &  &  & 100 & 8.9 $\pm$ 0.9 & 100 & 1.05 & 8.9 $\pm$ 0.9 & 100 & 1.20  \\
dermatology & 41 & 11.1 $\pm$ 1.2 & 95 & 7.5 $\pm$ 1.7 & 98.6 & 1.33 & 7.5 $\pm$ 1.7 & 98.6 & 0.00  \\
 &  &  & 90 & 7.4 $\pm$ 1.6 & 98.3 & 1.18 & 7.4 $\pm$ 1.6 & 98.3 & 0.00  \\
\midrule
 &  &  & 100 & 9.0 $\pm$ 1.1 & 100 & 2.85 & 9.0 $\pm$ 1.0 & 100 & 1.32  \\
kr-vs-kp & 86 & 10.9 $\pm$ 1.0 & 95 & 7.8 $\pm$ 2.0 & 97.6 & 5.16 & 7.5 $\pm$ 2.0 & 97.2 & 0.01  \\
 &  &  & 90 & 5.8 $\pm$ 1.7 & 91.6 & 4.92 & 5.6 $\pm$ 1.7 & 91.5 & 0.01  \\
\midrule
 &  &  & 100 & 8.9 $\pm$ 0.8 & 100 & 295.30 & 8.9 $\pm$ 0.8 & 100 & 1.36  \\
letter & 361 & 12.7 $\pm$ 1.4 & 95 & 8.7 $\pm$ 1.0 & 99.5 & 219.68 & 8.7 $\pm$ 0.9 & 99.6 & 0.01  \\
 &  &  & 90 & 8.4 $\pm$ 1.0 & 97.4 & 160.57 & 8.4 $\pm$ 0.9 & 97.5 & 0.01  \\
\midrule
 &  &  & 100 & 8.6 $\pm$ 1.1 & 100 & 5.50 & 8.6 $\pm$ 1.1 & 100 & 1.20  \\
soybean & 184 & 9.9 $\pm$ 1.7 & 95 & 8.5 $\pm$ 1.1 & 99.7 & 8.27 & 8.5 $\pm$ 1.1 & 99.7 & 0.00  \\
 &  &  & 90 & 7.9 $\pm$ 1.3 & 95.8 & 8.41 & 8.0 $\pm$ 1.3 & 96.3 & 0.00  \\
\midrule
 &  &  & 100 & 10.0 $\pm$ 0.3 & 100 & 6.56 & 10.1 $\pm$ 0.4 & 100 & 1.32  \\
spambase & 254 & 11.1 $\pm$ 0.5 & 95 & 5.1 $\pm$ 0.6 & 95.9 & 12.92 & 5.1 $\pm$ 0.8 & 96.8 & 0.01  \\
 &  &  & 90 & 3.1 $\pm$ 0.5 & 90.3 & 11.63 & 3.1 $\pm$ 0.6 & 90.4 & 0.02  \\
\midrule
 &  &  & 100 & 8.4 $\pm$ 0.7 & 100 & 13.80 & 8.4 $\pm$ 0.7 & 100 & 1.66  \\
texture & 175 & 8.8 $\pm$ 0.9 & 95 & 7.2 $\pm$ 1.0 & 98.9 & 18.02 & 7.3 $\pm$ 1.0 & 99.2 & 0.03  \\
 &  &  & 90 & 7.2 $\pm$ 0.9 & 98.5 & 31.23 & 7.2 $\pm$ 1.0 & 99.0 & 0.05  \\

\bottomrule[1.2pt]
\end{tabular}
}
\caption{ Assessing explanations of  $\mdrset$ and $\adrset$. 
	(For each dataset, we run the explainers on instances that obtain 
	AXp's (i.e.\  $\drset$ s.t. $\delta = 1$)) of length greater than 7)	
	Column {\bf \#I} denotes the number of tested instances. 
	%randomly 
	%picked from the training data. 
	%
	Column {$\bm\delta$} reports in (\%) the value of the threshold  $\delta$.
	{\bf Path}   reports the average 
	 depth of paths consistent with the instances. 
 	In column $\mdrset$ (resp.\ $\adrset$), column  {\bf Length} reports 
	the average length of the explanations, {\bf Precision} reports in (\%)  
	 the average precision (defined in (\ref{eq:drs})) of computed explanations 
	 and  column {\bf Time} reports in (sec)  the average runtime that takes 
	 the algorithm
	 to compute an explanation ($\pm$ denotes the standard deviation).
	%
	%{$\bm\drset$} shows the number in  (\%) of $\adrset$'s that are $\drset$'s. 
	%{\bf m$_{\bm\subseteq}$}	shows the number in  (\%) of $\adrset$'s that are 
	%subset-minimal, i.e.\  $\drset$'s. 
	 } 
\label{tab:paxp-res}
\end{table*}

\section{Experimental Results} \label{sec:res}
This section evaluates the algorithms proposed for computing $\mdrset$ 
and $\adrset$.
%Moreover, the assessment compares
%but it also 
The evaluation
includes a comparison  %of proposed algorithms 
with the %a state of the art
model-agnostic explainer Anchor~\cite{guestrin-aaai18}, aiming at
assessing not only the succinctness and precision of computed
explanations but also 
the scalability of our solution.

%\subsubsection{Prototype implementation.}
\paragraph{Prototype implementation.~}
A prototype implementation of the proposed
% algorithms is developed in Python. %as a Python script.
algorithms is developed in Python; %as a Python script.
whenever necessary, it instruments oracle calls to well-known SMT solver
z3\footnote{\url{https://github.com/Z3Prover/z3/}}~\cite{MouraB08} as
described in \cref{sec:rsdt}
\iftoggle{long}{%
  \footnote{%
    In this section, we equate the names $\wdrset$, $\adrset$, $\mdrset$
    and $\drset$ with their implementation using the algorithms
    described earlier in~\cref{sec:rsdt}.}.}{.}
Hence, the prototype implements the $\adrset$ procedure outlined 
in \cref{sec:adrset} and augmented  with a heuristic %preprocess
that orders the features in $\fml{X}$. The idea consists in computing 
the precision loss of the overapproximation of each $\fml{X}\setminus\{j\}$ 
%for each $j \in \fml{X}$ 
and then sorting the features from the less to the most important one. 
This strategy often allows obtaining the closest superset to a
$\drset$, in contrast  
to the simple lexicographic order applied over  $\fml{X}$.
(Recall that $\fml{X}$ is initialized to the set of features involved in the 
decision path.)   
Algorithm $\mdrset$ outlined in  \cref{sec:mdrset} implements
the two (multiplication- and addition-based) SMT encodings.
Nevertheless, preliminary results show that both encodings perform
similarly,
%somewhat similarly efficient,
with some exceptions where the addition-based encoding is much larger and so
slower. Therefore, the results reported below refer only to the
% non-linear encoding.  
multiplication-based encoding.  
%
% Furthermore,  $\mdrset$ is build upon the SMT solver z3~\cite{MouraB08}. 
% z3 is implemented in C++ and offers a Python package\footnote{\url{https://github.com/Z3Prover/z3/}}  
% as interface.
%Furthermore, the SMT solver z3~\cite{MouraB08} is used to instrument 
%the MaxSMT oracle calls. z3 is implemented in c++ and provides a python
%package\footnote{\url{https://github.com/Z3Prover/z3/}} that we import and 
%build upon our explainer.   

\paragraph{Experimental setup.~}
%\subsubsection{Experimental setup.}
%
The experiments are conducted on a MacBook Air with a 1.1GHz
Quad-Core Intel Core~i5  CPU with 16 GByte RAM running 
macOS  Monterey.
%
%Moreover, t
The benchmarks used in the experiments comprise publicly 
available and widely used datasets that originate from 
UCI ML Repository \cite{uci}.
%
%The number of training data (resp.\ features) in the target
%datasets varies from 12 to 57 (resp.\ 366 to 18668).
%
All the DTs are trained using the learning tool
\emph{IAI} (\emph{Interpretable AI})~\cite{bertsimas-ml17,iai}.
%
%The data are randomly split into 80\% and 20\% resp., of train and
%test set.
%
The maximum depth parameter in IAI is set to 16.
%%%To enforce IAI to produce accurate DTs, it is
%%%set to use the optimal tree classifier method with the maximal 
%%%depth of 16.
%
%Note that the training accuracy of all the models is above 89\%, 
%while the test accuracy varies from 64\% to 100\% and the tree
%depth of the trees varies from 12 to 16 and the total number of 
%nodes varies from 71 to 3261.
%%
%
As the baseline, we ran Anchor with the default explanation precision
of 0.95. Two assessments are performed with Anchor: (i) with the original  
training data\footnote{The same training set used to learn the model.} 
that follows the data distribution; (ii) with 
using sampled data that follows a uniform distribution. 
Our setup assumes that all instances of the feature space 
%$\mbb F$ 
are possible, and so there is no assumed probability distribution 
over the features. %(i.e. uniformly distributed data).  
Therefore in order to be fair with Anchor, we further assess 
Anchor with uniformly sampled data. 
%
%The samples are generated using \texttt{numpy} package, 
%so that each feature is sampled individualy to generate 
%one sample and then labeled with the learnt DT. 
%The number of generated samples equals to the number of training 
%data, i.e. 80\% of the dataset. 
%
(Also, we point out that the implementation of Anchor demonstrates that   
 it can generate samples that do not belong to the input distribution. 
 Thus,  there is no guarantee that these samples come from the input
 distribution.)  
Also, the prototype implementation was tested with varying  the
threshold  $\delta$ while Anchor runs guided by its own metric.

\begin{comment}
Besides, the prototype implementation was run for the values of $\delta$ from
$\{\textrm{0.9},   \textrm{0.95}, \textrm{1.0}\}$.
%
It should be observed that the proposed experiment gives an advantage
to Anchor, as Anchor is allowed to computes explanations guided by its
own metric, whereas $\adrset$ and $\mdrset$ \emph{know nothing} about
this metric (which they will be assessed with).
%
\end{comment}

%Finally, the experiments are conducted on a MacBook Air with a 1.1GHz
%Quad-Core Intel Core~i5  CPU with 16 GByte RAM running macOS X
%Monterey.

%\subsubsection{Results.}
\paragraph{Results.~}
\cref{tab:res} summarizes the results of our experiments. 
One can observe that $\mdrset$ and $\adrset$ compute succinct 
explanations (i.e. of average size $7\pm2$ \cite{miller-pr56}), for 
the majority of tested instances across all datasets, 
noticeably shorter than consistent-path explanations.  
More importantly, the computed explanations are trustworthy and 
show good quality precision, e.g.\ {\it dermatology}, {\it soybean} 
and {\it texture} show avg.\ precisions greater than 98\% for all values
of $\delta$. 
Additionally, the results clearly demonstrate that our proposed SMT 
encoding scales for deep DTs with runtimes on avg.\ less than 20 sec 
for the largest encodings while the runtimes of $\adrset$ are negligible,  
never exceeding 0.01 sec. 
Also, observe that the lion's share of over-approximations computed by
$\adrset$ are often subset-minimal $\drset$'s, mostly as short as
computed $\mdrset$'s.  
This demonstrates empirically the advantage of the algorithm, i.e.\
%
% Thus, the results %demonstrate the effectiveness of the algorithm
% showcase the  empirical advantages of the algorithm. 
%
% Notably,
%
%
in practice one may rely on the computation of $\adrset$'s, which pays
off in terms of (1)~performance, (2)~sufficiently high probabilistic
guarantees of precision, and (3)~good quality over-approximation of
subset-minimal $\drset$'s.
% 
% one may opt for approximating subset-minimal (rather than smallest-size) 
% but not minimum size,
% explanations as they offer suffiently high probabilistic guarantees 
% of precision.
%
In contrast, Anchor is unable to provide precise and succinct
explanations in both settings of data and uniform distribution.
Moreover, we observe 
that Anchor's explanations often include features that are not involved 
in the consistent path, e.g.\  for {\it texture} less than 20\% of an explanation  
is shared with the consistent path.
%, which might be problematic in some situations. 
%(This observation was also pointed out in a recent 
%work  about explaining XGBoost models~\cite{ignatiev-ijcai20}.) 
(This trend was also pointed out by~\cite{ignatiev-ijcai20}.)
In terms of average  runtime, Anchor is overall slower, being
outperformed by the computation of $\adrset$ by several orders of
magnitude.

Focusing solely on large $\drset$'s of size greater than 7, \cref{tab:paxp-res} 
reports the detailed results of $\adrset$ and $\mdrset$ tested with probabilities 
$\delta \in \{0.9, 0.95, 1.0\}$. 
(Note that these results are included in summarized results in 
\cref{tab:res} shown in the paper.)
Hence, the purpose is to show the empirical advantages 
 of our algorithms for computing provably succinct explanations 
when the instances to explain match very large tree paths 
and AXp's are also large 
(exceeding the cognitive limits of human decision makers~\cite{miller-pr56}). 
As can be seen from the table, both methods deliver shorter 
explanations, with allowing to drop a small threshold probability 
of the precision (i.e. $1 - \delta$).
For example, the average size of computed $\drset$'s  in {\it spambase} 
decreases from 10 features to 5 features while respecting  a precision 
greater than 0.95.

Overall, the experiments demonstrate that our approach
efficiently computes succinct and provably precise explanations for
large DTs.
The results also substantiate the limitations of model-agnostic
explainers, both in terms of explanation quality and computation
time.
%low-quality explanations as well as poor performance exhibited by
%Anchor, thus suggesting that model-agnostic approaches do not offer a
%viable alternative.

%%\input{relw}
\section{Conclusions} \label{sec:conc}

Abductive explanations are guaranteed to be sufficient for the
prediction while being
subset-minimal~\cite{darwiche-ijcai18,inms-aaai19,darwiche-ecai20}.
However, the size of abductive explanations may be beyond the
cognitive reach of human decision makers~\cite{miller-pr56}.
Recent work proposed $\delta$-relevant sets%
%approximate explanations with rigorous probabilistic guarantees
~\cite{kutyniok-jair21}.
%i.e.\ $\delta$-relevant sets.
The downside of this earlier work is the practically prohibitive
computational complexity of deciding whether a set of features is such
an approximate explanation.
Building on recent work on explaining decision
trees~\cite{barcelo-nips20,hiims-kr21,marquis-kr21}, this paper shows
that computing a smallest $\wdrset$ (which generalizes smallest 
$\delta$-relevant sets) can be solved with a logarithmic number of
calls to an NP oracle in the concrete case of DTs. Furthermore, the
paper argues that existing algorithms for finding subset-minimal sets
will yield tight over-approximations ($\adrset$'s) of subset-minimal
sets when used for refining a given $\wdrset$.
The experimental results demonstrate that the proposed SMT encodings
scale in practice, and that the computation of $\adrset$'s most often
yields $\wdrset$'s that are indeed subset-minimal, i.e.\ a $\drset$.
Furthermore, the paper offers additional evidence to the poor quality
of explanations computed by state of the art model-agnostic
explainers~\cite{guestrin-aaai18}.

\iftoggle{long}{%
  A possible limitation of our work is that, albeit the approach is 
  model-accurate, input distributions are not accounted for. (This
  observation applies to past work on abductive explanations.)
  %(..., with the exception of upcoming work~\cite{rubin-aaai22}.)
  There is recent work on constraining the inputs of classifiers in
  order to take input constraints into
  account~\cite{rubin-aaai22}. Integration of the two lines of work is
  expected to provide additional rigor in computing approximate
  explanations.}{}

\subsubsection*{Acknowledgments}
%\paragraph{Acknowledgments.}
  %
  This work was supported by the AI Interdisciplinary Institute ANITI, 
  funded by the French program ``Investing for the Future -- PIA3''
  under Grant agreement no.\ ANR-19-PI3A-0004, and by the H2020-ICT38
  project COALA ``Cognitive Assisted agile manufacturing for a Labor
  force supported by trustworthy Artificial intelligence''.
  %

% RequiredL: \usepackage{etoolbox}
%\providetoggle{mkbbl}
\newtoggle{mkbbl}
% Contents if using bibtex: "\settoggle{mkbbl}{true}"
% Contents if inputing pre-generated file: "\settoggle{mkbbl}{false}"

\settoggle{mkbbl}{false}
 % file is automatically generated

% For arxix paper production, and since arXiv does not allow for
% bibtex, we need to create a .bbl file to include upon submission
% to arXiv.
\iftoggle{mkbbl}{
  \bibliographystyle{named}
  \bibliography{refs,dts,team}
}{
  % Import bibl (original .bbl) file

}
%

%\appendix
%\input{appendix}

\end{document}